\documentclass[runningheads]{llncs}
\usepackage{graphicx}
\usepackage{multirow}
\begin{document}
\title{Unsupervised Domain Adaptation for Mammogram Image Classification: A Promising Tool for Model Generalization}

%
%
\author{Yu Zhang\inst{1}* \and
Gongbo Liang\inst{1} \and
Nathan Jacobs\inst{1}\and
Xiaoqin Wang\inst{2}}
\authorrunning{Y. Zhang et al.}
%
\institute{Department of Computer Science, University of Kentucky, USA \and
Department of Radiology, University of Kentucky, USA\\
\email{Email: y.zhang@uky.edu*}}
%
\maketitle              
\begin{abstract}
Generalization is one of the key challenges in the clinical validation and application of deep learning models to medical images. Studies have shown that such models trained on publicly available datasets often do not work well on real-world clinical data due to the differences in patient population and image device configurations. Also, manually annotating clinical images is expensive. In this work, we propose an unsupervised domain adaptation (UDA) method using Cycle-GAN to improve the model’s generalization ability without using any additional manual annotations.

\keywords{Breast cancer  \and Mammogram \and Deep learning \and Unsupervised domain adaptation \and GAN.}
\end{abstract}
\section{Hypothesis}
We know if we train a deep learning model on a labeled dataset A (source domain), it may achieve high performance on A but low performance on an unlabeled dataset B (target domain) because A and B may have different attributes. We hypothesize the UDA method will improve the model’s performance on B while maintaining the high performance on A.

\section{Methods}
\subsection{Dataset}
The public mammogram dataset Digital Database of Screening Mammography (DDSM)~\cite{heath1998current} and a private mammogram dataset, UKY~\cite{wang2020inconsistent}, are used in this work. These two datasets have different attributes: DDSM contains digitalized screen film mammograms and UKY contains full-field digital mammograms recently collected from a comprehensive breast imaging center. Several recent works explored the different attributes of those two or other similar datasets~\cite{zhang20192d,liang2019joint,whole}. In this work, we use 1860 positive and 2781 negative images from DDSM and 1922 positive and 2330 negative images from UKY. We split the data in 80\% for training and 20\% for testing.

\begin{figure*}
  \centering
  \includegraphics[width=1.0\textwidth]{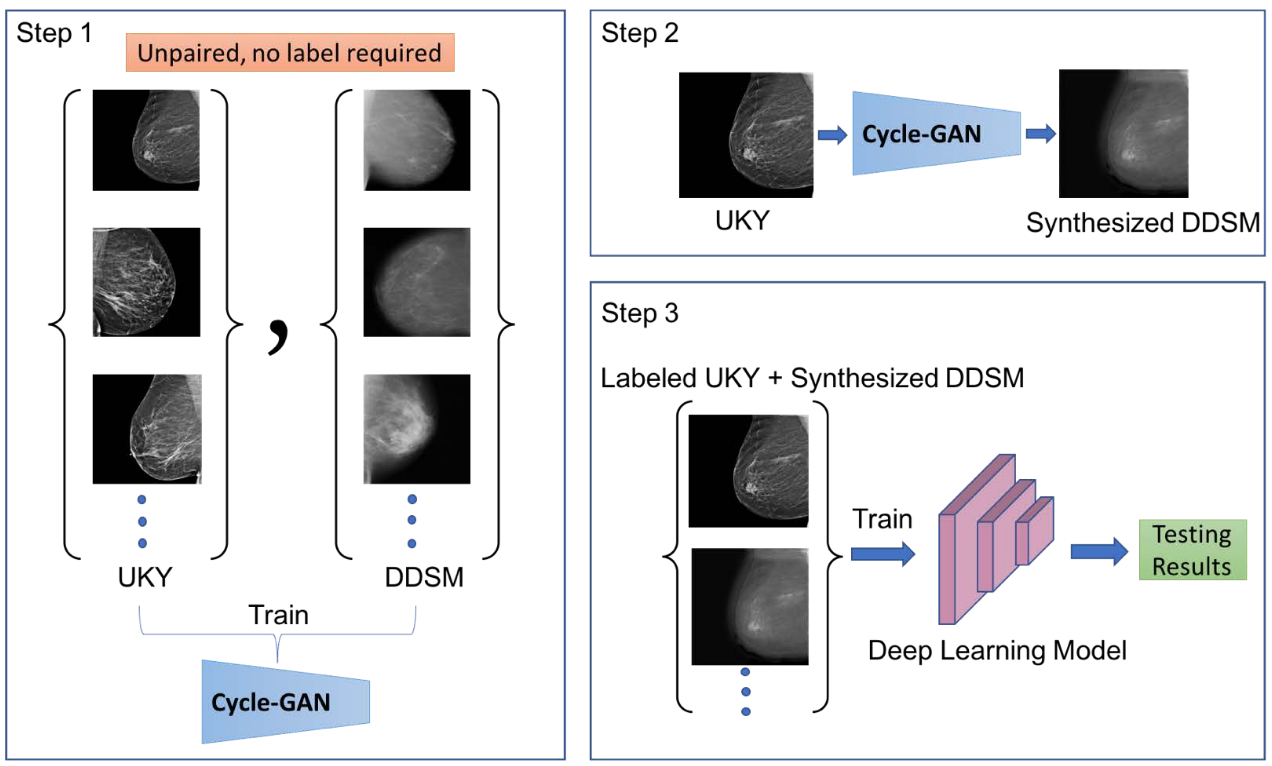}
  \caption{Stepwise illustration of our unsupervised domain adaptation(UDA) method. Step 1) train Cycle-GAN by using unpaired, unlabeled UKY and DDSM datasets; Step 2) translate UKY to DDSM; 3) train deep learning models by using UKY and synthesized DDSM.}
  \label{figs:main}
\end{figure*}

\subsection{Method} Figure~\ref{figs:main} illustrates our UDA method. We first train the Cycle-GAN~\cite{zhu2017unpaired} on unpaired images without any labels, then we synthesize DDSM data from UKY data to generate training samples in the target domain. Finally, we train a deep neural network on a mixture of UKY and synthesized DDSM images. We compared our UDA method with the baseline method, which trains on one dataset and directly tests on another dataset. In addition, we train the models on labeled DDSM and synthesized UKY by switching the source and target domains for a two-way verification.

\section{Experimental Results}
Our results are summarized in Table~\ref{tab1}. Two off-the-shelf architectures are used for evaluation: AlexNet~\cite{krizhevsky2012imagenet} and ResNet~\cite{he2016deep}. When training and testing on different datasets, UDA achieves significant improvement compared to the baseline. For instance, when we trained on UKY and tested on DDSM with AlexNet, the baseline only achieved 0.516 auROC while UDA achieved 0.601 auROC. The table also shows when training and testing on the same dataset, UDA maintains similarly high performance, which verified our hypothesis. 

\begin{table}[]
\caption{Testing Results of Different Methods.}\label{tab1}
\begin{tabular}{|c|c|c|c|c|c|}
\hline
\multirow{3}{*}{Training Set} & \multirow{3}{*}{Testing Set} & \multicolumn{4}{c|}{Mean auROC (95\% Confidence Interval)}              \\ \cline{3-6} 
                              &                              & \multicolumn{2}{c|}{AlexNet}       & \multicolumn{2}{c|}{ResNet50}      \\ \cline{3-6} 
                              &                              & Baseline    & UDA                  & Baseline    & UDA                  \\ \hline
\multirow{2}{*}{UKY}          & DDSM                         & ${0.516 \pm 0.004}$ & $\bf{0.601 \pm 0.005}$ & ${0.624 \pm 0.004}$ & $\bf{0.672 \pm 0.002}$ \\ \cline{2-6} 
                              & UKY                          & ${0.785 \pm 0.003}$ & ${0.769 \pm 0.007}$          & ${0.836 \pm 0.008}$ & ${0.869 \pm 0.016}$          \\ \hline
\multirow{2}{*}{DDSM}         & UKY                          & ${0.491 \pm 0.007}$ & $\bf{0.578 \pm 0.002}$ & ${0.565 \pm 0.002}$ & $\bf{0.674 \pm 0.003}$ \\ \cline{2-6} 
                              & DDSM                         & ${0.673 \pm 0.015}$ & ${0.653 \pm 0.024}$          & ${0.762 \pm 0.008}$ & ${0.759 \pm 0.012}$          \\ \hline
\end{tabular}
\end{table}


\section{Conclusion}
Our results show that the proposed UDA method improves deep learning models’ generalization without requiring expensive manual annotations. However, there is still room for improvement. We expect combining improved versions of Cycle-GAN with small amounts of labeled data in the target domain will help bridge the gap.

\section{Statement of Impact}
Despite the reported high performance of deep learning models in crafted training data, generalization remains the challenge due to differences in publicly available and real-world clinical datasets. Our UDA method helps train models that can generalize between datasets, thereby significantly improving the results and lowering the cost of using deep learning models in clinical practice.

%
%
%
\bibliographystyle{splncs04}
%
\bibliography{cbib.bib}





\end{document}